\documentclass[10pt,twocolumn,letterpaper]{article}

\usepackage{iccv}
\usepackage{times}
\usepackage{epsfig}
\usepackage{graphicx}
\usepackage{amsmath}
\usepackage{amssymb}

\usepackage{enumerate}
\usepackage{bigstrut}
\usepackage{subcaption}
\usepackage{bm}
\usepackage{algorithm,lipsum}
\usepackage{wrapfig}
\usepackage[noend]{algpseudocode}
\newcommand{\myparagraph}[1]{\vspace{6pt}\noindent{\bf #1}}
\usepackage[pagebackref=true,breaklinks=true,letterpaper=true,colorlinks,bookmarks=false]{hyperref}

 \iccvfinalcopy % *** Uncomment this line for the final submission

\def\iccvPaperID{10536} % *** Enter the ICCV Paper ID here

\ificcvfinal\fi

\begin{document}
	
	%%%%%%%%% TITLE
%	\title{Tf-GCZSL: Task-Free Generalized Continual Zero-Shot Learning}
	\title{Online Lifelong Generalized Zero-Shot Learning}
		
		\author{Chandan Gautam\\
	Indian Institute of Science, Bangalore\\
	%	Institution1 address\\
	{\tt\small  chandang@iisc.ac.in}
	\and
	Sethupathy Parameswaran\\
	Indian Institute of Science, Bangalore\\
	{\tt\small sethupathyp@iisc.ac.in}
	\and
	Ashish Mishra\\
	IIT Madras\\
	{\tt\small mishra@cse.iitm.ac.in}
	\and
	Suresh Sundaram\\
	Indian Institute of Science, Bangalore\\
	{\tt\small vssuresh@iisc.ac.in}
}
	
	\maketitle
	\ificcvfinal\thispagestyle{empty}\fi
	
	%%%%%%%%% ABSTRACT
	\begin{abstract}
		Methods proposed in the literature for zero-shot learning (ZSL) are typically suitable for offline learning and cannot continually learn from sequential streaming data. The sequential data comes in the form of tasks during training. Recently, a few attempts have been made to handle this issue and develop continual ZSL (CZSL) methods. However, these CZSL methods require clear task-boundary information between the tasks during training, which is not practically possible. This paper proposes a task-free (i.e., task-agnostic) CZSL method, which does not require any task information during continual learning. The proposed task-free CZSL method employs a variational autoencoder (VAE) for performing ZSL. To develop the CZSL method, we combine the concept of experience replay with knowledge distillation and regularization. Here, knowledge distillation is performed using the training sample's dark knowledge, which essentially helps overcome the catastrophic forgetting issue.
		Further, it is enabled for task-free learning using short-term memory. Finally, a classifier is trained on the synthetic features generated at the latent space of the VAE. Moreover,	the experiments are conducted in a challenging and practical ZSL setup, i.e., generalized ZSL (GZSL). These experiments are conducted for two kinds of single-head continual learning settings:  (i) mild setting-: task-boundary is known only during training but not during testing; (ii) strict setting-: task-boundary is not known at training, as well as testing. Experimental results on five benchmark datasets exhibit the validity of the approach for CZSL.
	\end{abstract}
	
	%%%%%%%%% BODY TEXT
	\section{Introduction}
	Conventional supervised machine learning (ML)  and, more recently, deep learning algorithms have shown remarkable performance on various tasks (e.g., classification/recognition) in various domains, such as Computer Vision and Natural Language Processing \cite{ResNet2015,alexnet}.  Despite the recent success of supervised ML/deep learning algorithms, they have two crucial limitations.
	\begin{enumerate}
		\item Conventional machine learning models have restricted themselves to training classes. When any example from the novel/unseen class exists in the test set, such models can not be appropriately categorized.
		\item Conventional machine learning models cannot continually learn over time by accommodating new knowledge while retaining previously learned experiences due to catastrophic forgetting.
	\end{enumerate}
	Recently, the first limitation is addressed by zero-shot learning (ZSL), where we classified objects from classes that are not available at training time \cite{ZSL_back,ZSL-region,ZSL_side,gzsl1}. The second limitation is resolved by continual/lifelong learning \cite{kirkpatrick2017overcoming,shin2017continual,chaudhry2018riemannian}. A traditional ZSL approaches having a problem for sequential training, and continual learning approaches can not handle unseen class objects. Therefore, a more preferable and desirable approach needs to tackle sequential training and unseen object problems simultaneously. This paper aims to leverage the advantages of both zero-shot learning and continual learning in a single framework. %Consequently, we propose a single model that can predict the unseen class object and adapt to a new task without forgetting the knowledge about the previous task.
	
	Zero-shot learning is an interesting paradigm to classify objects from classes that are not available at training time. Zero-shot learning (ZSL) methods have attracted considerable attention in recent years owing to their ability to classify unseen/novel class examples. Earlier approaches for zero-shot learning are based on the embedding function between visual and semantic space and are therefore biased towards the seen classes. Generative models recently address this issue, where the generative models are used to synthesize visual features directly from semantic class descriptors. Feature generative methods provide a shortcut to cast the zero-shot learning problem into a conventional classification problem \cite{f-CLSWGAN,f-VAEGAN,CVAE,GFZSL,vinay_meta,episodecvpr2020}.
	
	The conventional training approach of ZSL trains the model on different classes (of the same dataset) under the assumption that all of the concerned data is available together during training. However, this is not always the case. There can be instances when new data arrives in a stream for training. The samples available in the newly arrived streaming data might belong to the existing or newly discovered classes, which need to be added to the dataset for updating the knowledge of the ZSL model. If we do not update the ZSL model on newly arrived data, then the model's prediction might be wrong as its knowledge is not updated. One way of handling this issue is to train the model from scratch with the inclusion of each incoming sample into the dataset every time. However, it will be a tedious and computationally expensive task to train each incoming sample model. Moreover, we need to store both previous and current data for training the ZSL model, which is not a feasible option due to memory constraints. 
	%which is not always possible as access to private data is restricted.
	Continual/incremental/lifelong learning can address the above concern by enabling the training of the ZSL model in a sequential manner with the preservation of accumulated (previous) knowledge while acquiring the new knowledge \cite{shin2017continual,lopez2017gradient,chaudhry2019tiny}. Consequently, we propose a single model that can predict the unseen class object and adapt to a new task without forgetting the knowledge about the previous task. This model is known as continual zero-shot learning (CZSL). It can update its current knowledge continuously without forgetting previous information in contrast to the conventional ZSL approaches. The CZSL method is a broad generalization of zero-shot learning. %The model is trained sequentially in seen class examples across the different datasets without forgetting the previous knowledge. 
	A few CZSL methods have been proposed yet for this setup \cite{wei2020lifelong,skorokhodov2020normalization}. Both existing works require task-boundary information during training. However, we are the first to propose CZSL for a task-free learning/task-free setting, which is a strict single-head setting.  LZSL \cite{wei2020lifelong} considers one whole dataset as a task and trains a separate attribute encoder-decoder for each task (dataset); therefore, it is a very trivial setting and does not support class-incremental setup for continual learning. More importantly, it requires the task boundary information during testing; therefore, this setup is not suitable for a single-head setting.

	%\textcolor{red}{Here we need to explain some work for CZSL and there shortcoming and how our model addresses them}
	
	The contributions of our proposed approach are summarized as:
	\begin{enumerate}
		\item  To the best of our knowledge, this is the first work that proposes continual zero-shot learning for the task-free setting. The existing approaches \cite{wei2020lifelong,skorokhodov2020normalization} are only compatible when task-boundary is either present during training or both training and testing.
		\item This paper also provides the novel evaluation setting for CZSL as the existing settings \cite{wei2020lifelong,skorokhodov2020normalization} are not suitable for task-free learning.   
		\item To enable the generative model for CZSL, the proposed approach employs experience replay with knowledge distillation. Here, we don't use the student-teacher network strategy for knowledge distillation (KD). Instead of that, we store the required information in the memory of the corresponding sample to perform KD. The stored information is generally known as dark knowledge \cite{hinton2014dark}.  
		\item To enable the model for task-free learning, this paper proposes short-term memory-based two different task-free learning strategies, which are compatible with any ZSL method.
		\item An extensive experimental results validate the effectiveness of the proposed task-free continual ZSL method.   
	\end{enumerate}

	\section{Related Work}
	\subsection{Zero-shot learning}
	Recently, ZSL has attracted considerable attention due to handling unknown objects during testing. It transfers knowledge from seen classes to unseen classes via class attributes. Earlier proposed approaches for ZSL primarily were discriminative or non-generative (i.e., embedding-based) in nature \cite{akata2016label,ConSE, xian2016latent, LampertNH14, zhang2015zero, socher2013zero, CSSL1, CSSL2, CSSL3, CSSL4, CSSL5}. Non-generative methods learn an embedding from visual space to semantic space or vise versa via a linear compatibility function \cite{akata2016label,ConSE, xian2016latent, LampertNH14}. These approaches are based on the assumption that the class attributes of seen and unseen classes share many similarities. Embedding-based approaches represent image class as a point hence unable to capture intra-class variability.
	
	The generalized ZSL (GZSL) problem is potentially more practical and challenging where the training and the test classes are not disjoint. As ZSL models train using seen class examples, most of the embedding-based approaches show a strong bias towards the seen classes in GZSL. VAE and GAN are the backbones for generative models used to synthesize the examples for several applications. The ability to synthesize the seen/unseen class examples from class attributes using VAE/GAN is the basis of generative models-based ZSL.
	Generative models have recently shown promising results for both ZSL and GZSL setups using synthesized examples for unseen classes to train a supervised classifier \cite{vermageneralized,cycle-UWGAN, f-CLSWGAN, DGAN, GZSL_few, ZSL_back, ZSL_side, f-VAEGAN, schonfeld2019generalized}.  A particular advantage of the generative models is that they transform a ZSL problem into a typical supervised learning problem using the synthesized examples of unseen classes to train a supervised classifier. 
	
	\subsection{Continual Learning}
	
	Continual learning learns from streaming data with two objectives: avoid catastrophic forgetting (preserve experience while learning on new tasks) and avoid intransigence (update new knowledge and transfer the previous knowledge). The whole work of continual learning can be broadly categorized into two parts: (i) regularization-based methods \cite{kirkpatrick2017overcoming,rebuffi2017icarl,chaudhry2018riemannian} (ii) replay-based methods \cite{lopez2017gradient,shin2017continual,chaudhry2018efficient,hayes2019memory,chaudhry2019tiny}. Most of the continual learning works are focused on multi-head setting \cite{kirkpatrick2017overcoming,chaudhry2018riemannian,rebuffi2017icarl}. In recent years, task-free learning receives a surge of interest among researchers \cite{aljundi2019task,aljundi2019gradient,darkneurips2020,gmedicml2020} as it is a more practical continual learning setting than a multi-head setting. This paper is focused on experience replay for task-free learning. 
	
	%Generally, continual learning tests under two types of evaluation settings \cite{chaudhry2018riemannian}: (i) single-head: task identity is not known during and multi-head. 
	
	\subsection{Continual Zero-shot Learning} 
	In a traditional continual learning setting, training and testing data contain the same number of classes for classification. However, in the CZSL setting, training data also contains some unseen classes with their description in textual form, and a classifier should be able to classify these unseen classes during testing. Most recently, CZSL \cite{wei2020lifelong,skorokhodov2020normalization} has drawn increasing
	interest. To the best of our knowledge, only a handful
	the number of work is available for this problem. Chaudhry et al. \cite{chaudhry2018efficient} develop an average gradient episodic memory (A-GEM) -based CZSL method for a multi-head setting. A generative model-based CZSL \cite{wei2020lifelong} method is also developed for multi-head setting. Most recently, Skorokhodov and Elhoseiny \cite{skorokhodov2020normalization} develop an A-GEM-based CZSL method for a single-head setting; however,,, it is not a strict single-head setting as task-identity is required during training. This paper develops a CZSL method for a strict single-head setting where task identity is neither known during training nor during testing.

	% \section{Related Work of CZSL}
	%More importantly, combining continual learning with zero-shot learning enables our model to train sequentially for a different task without forgetting previous knowledge. Our CZSL model can handle unknown class objects and adopt in various tasks without missing the previous tasks. We describe the various version of the CZSL framework in the next subsections.

	\section{Problem Formulation}
	Formally, a CZSL is divided among $T$ tasks ($t\in{1,\hdots,T}$), where each $t^{th}$ task consists of training and testing data stream. Generally, the training stream $\mathcal{D}^t_{tr}$ for $t^{th}$ task contains only the information of seen classes, which consists of feature vector $x_i^t$, task identity  $\iota_i^t$ (it provides task-boundary), class label $y_i^t$, and class attribute information $a_i^t$. Where $i$ represents $i^{th}$ sample from the whole training samples $n_{tr}$ of $t^{th}$ task.  In addition, training stream also contains class attribute information for unseen classes, i.e., $\mathcal{U_C} = \{(a_i)_{i=1}^{n_{uc}}\}$ where $n_{uc}$ denotes number of unseen classes. This is the key information which enables model for performing CZSL. Similarly, testing stream $\mathcal{D}^t_{ts}$ consists of $\{(x_i^t, \iota_i^t, y_i^t)_{i=1}^{n_{ts}}\}$, where $n_{ts}$ is total number of test samples for $t^{th}$ task. Here, testing class label is only used for evaluation purpose. In this paper, we address single-head setting for two possible situations: (i) \textbf{task-agnostic prediction}: when task boundary is only available during training but not in testing, i..e., $\mathcal{D}^t_{tr} = \{(x_i^t, \iota_i^t, y_i^t, a_i^t)_{i=1}^{n_{tr}}\}$ and $\mathcal{D}^t_{ts} = \{(x_i^t, y_i^t)_{i=1}^{n_{ts}}\}$; (ii) \textbf{task-free learning}: when task boundary is neither available during training nor testing, i.e., $\mathcal{D}_{tr}^{t} = \{(x_i, y_i, a_i)_{i=1}^{n_{tr}}\}$ and  $\mathcal{D}_{ts}^{t} = \{(x_i, y_i)_{i=1}^{n_{ts}}\}$.
	
	\section{Task-Free Generalized Continual Zero-shot Learning: Tf-GCZSL}
	
	%\noindent {\bf Problem Formulation}:
	
	\begin{figure*}[h!]
		\centering
		\includegraphics[width=1.0\linewidth]{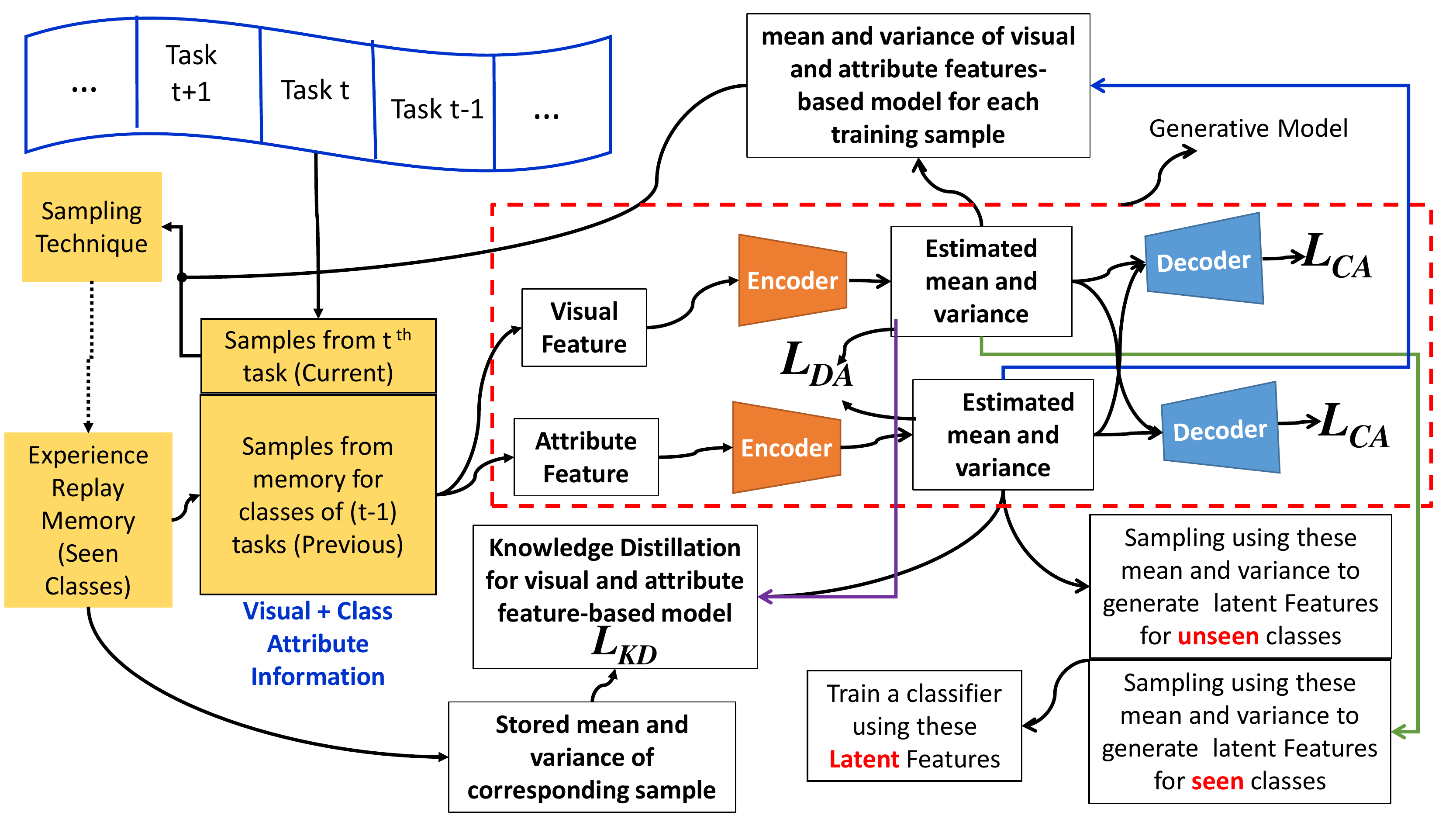}
		\caption{Proposed Task-free-GCZSL Framework. Here, task is just shown for better understanding, however, it is not required either during training or testing as reservoir sampling technique does not depend on the task information. We do not show the short-term memory  in the diagram for sake of simplicity, however, it is described in the Algorithm \ref{alg:ALG1} and \ref{alg:ALG2}. Actually, incoming samples come to the short-term memory before training in the proposed method. Although, this framework is still a task-free framework without that short-term memory, however, performance significantly enhances if we use short-term memory}
		\label{fig:Tf-GCZSL}
	\end{figure*}
	
	%\subsection{Tf-GCZSL}
	Earlier developed methods for CZSL contains task-boundary either during training (task-agnostic prediction setting) \cite{skorokhodov2020normalization} or during training and testing both (multi-head setting) \cite{chaudhry2018efficient,wei2020lifelong}. In this section, a task-free continual learning method is proposed for the GZSL framework, i.e., task-free generalized continual zero-shot learning (Tf-GCZSL). Tf-GCZSL is developed based on the concept of experience replay (ER) with knowledge distillation (KD) and regularization. Here, knowledge distillation is performed by using the dark knowledge \cite{hinton2014dark} instead of using the teacher network (i.e., teacher network is the immediate previous network in the case continual learning). Dark knowledge is the soft labels of the training samples of the previous tasks \cite{hinton2014dark}. Training on these soft labels helps in alleviating the catastrophic forgetting and regularize the model for better performance. The proposed Tf-GCZSL method (as shown in Figure \ref{fig:Tf-GCZSL}) use latent space (i.e., output of the encoder) information as a dark knowledge for performing KD. In Tf-GCZSL, A generatic ZSL method, CADA \cite{schonfeld2019generalized}, is used as a base method due to it's good performance for the ZSL in the current literature; however, any ZSL method can be used instead of CADA as the proposed Task-free GCZSL framework isa generic. It minimizes 3 kinds of losses:    
	
	\myparagraph{VAE loss}: It minimizes two standard VAE losses simultaneously for feature and attribute encoder-decoder network: Kullback–Leibler (KL) divergence \cite{kullback1951information} loss ($\mathcal{L}_{KL}$) and reconstruction loss ($\mathcal{L}_{Re}$).
	
	\myparagraph{Distribution-alignment loss (DA)}: It minimizes the distribution between the latent space information of feature and attribute encoder. 
	\begin{equation}
	\mathcal{L}_{DA} = (\|\mu_{Af} - \mu_{Vf}\|_2^2 + \|(\Sigma_{Af})^{\frac{1}{2}} - (\Sigma_{Vf})^{\frac{1}{2}}\|_{\mathcal{F}}^2)^{\frac{1}{2}},
	\end{equation}
	where $\mu_{Vf}$ and $\Sigma_{Vf}$ are the estimated mean and variance by visual encoder $E_{Vf}$, respectively. $\mu_{Af}$ and $\Sigma_{Af}$ are the estimated mean and variance by attribute encoder $E_{Af}$, respectively, and $\mathcal{F}$ represents Frobenius norm. 
	
	\myparagraph{Cross-alignment loss (CA)}: It is cross-reconstruction loss between the output of feature and attribute decoder and is given as
	\begin{equation}
	\mathcal{L}_{CA} = |a - D_{At}(E_{Vf}(x))| + |x - D_{Vf}(E_{Af}(a))|,
	\end{equation}
	where $x$, $a$, $D_{Af}$, and $D_{Vf}$ denote visual feature vector, class attribute vector, attribute decoder, and visual decoder, respectively. 
	
	\noindent The overall loss ($\mathcal{L}_{G}$) of a generative method (i.e., CADA) for performing ZSL is as follows:
	\begin{equation}\label{cada_loss}
	\mathcal{L}_{G} = \mathcal{L}_{Re} + \beta\mathcal{L}_{KL} + \gamma \mathcal{L}_{CA} + \delta \mathcal{L}_{DA},
	\end{equation}
	where $\beta$, $\gamma$ and $\delta$ are the weighting factors.
	
	\subsection{Experience Replay and Task-free Strategies for CZSL}
	Experience Replay (ER) is a well-known method to alleviate catastrophic forgetting in continual learning framework for handling general classification task. However, in this paper, we combine ER and knowledge distillation for task-free generalized continual zero-shot learning (Tf-GCZSL). In Tf-GCZSL, like an experience replay strategy, ER stores the previously learned samples in a small memory $\mathcal{M}$ and replays it later for training the model. A model is jointly trained by the samples from the replay memory $\mathcal{M}$ and the samples from the current streaming data. This joint training helps the model in retaining the past knowledge. Here, we need to address two important issues: (i) replay memory capacity $\mathcal{M}$ is full and (ii) task-free setting during training. In order to handle full memory capacity, we employ a task-independent sampling technique like reservoir sampling, i.e., task-boundary is not known in the streaming data. Reservoir sampling selects $\mathcal{M}$ random samples from the streaming data with probability $\frac{\mathcal{M}}{l}$. Here, $l$ is the number of samples seen so far and it is not required to known in advance. Further, CZSL model needs to train in the task-free setting. In this setting, sample arrives one-by-one to the model for training, and train the model by a single sample heavily overfit the CZSL model. As the task-boundary is not known, it is difficult to optimize the model parameters and determining stopping criteria. To handle this issue, we propose two different task-free learning strategies using short-term memory. One should noted that the short-term memory is different from memory present in experience replay. ($\mathcal{M}$): %(i) one epoch training (ii) short-term training. 
	
	\myparagraph{(i) task-free CZSL strategy-1:}  When the memory $\mathcal{M}$ reaches the maximum capacity first time, we stop the incoming data stream for a while and optimize the model once. After completing this one-time optimization, training data stream resumes with very small-sized short-term batch memory ($\mathcal{M}_b$) to store the incoming data stream. This short-term memory is simply a very small batch which is required to pass only once to the model for training without multiple epochs. After completing the training using  $\mathcal{M}_b$, the memory is cleared to store the sample from incoming data stream. The process is repeated until all samples from stream of training data are presented to the model. Since, this strategy does not require multiple epochs, it is really fast in learning the samples. It is referred as Tf-GCZSL$_{\mathcal{M}_b}$. The pseudocode of this procedure is provided in Algorithm \ref{alg:ALG1}.                
	
	\myparagraph{(ii) task-free CZSL strategy-2:} In this strategy, we employ a larger short-term memory, i.e., $\mathcal{M}_{st}$ is larger than $\mathcal{M}_b$. The incoming samples are stored in $\mathcal{M}_{st}$ until it becomes full. Once, this memory becomes full then we stop the incoming training samples for a while and train the model for multiple epochs for better generalization. After completing the training using $\mathcal{M}_{st}$, the $\mathcal{M}_{st}$ is cleared to store the samples from the incoming data stream. The process is repeated untill there is no samples from data stream. Tf-GCZSL with this strategy is referred as Tf-GCZSL$_{\mathcal{M}_{st}}$. The pseudocode of this procedure is provided in Algorithm \ref{alg:ALG2}.   
	
	\begin{algorithm}[!h]
		\caption{{\bf Task-free Learning Strategy-1}}
		\textbf{Input:} \text{Data stream $\mathcal{D}_{tr}$, Short-term memory $\mathcal{M}_{b}$}\\
		\textbf{Output:} \text{Trained model}
		\begin{algorithmic}[1]
			\State $optimization\_done$ $\gets$ False
			\For{$i^{th}$ sample in $\mathcal{D}_{tr}$}
			\State Store the incoming $i^{th}$ sample in replay memory $\mathcal{M}$ using Reservoir sampling strategey
			\If{Replay memory $\mathcal{M}$ is full and $optimization\_done$ is False}
			\State Stop the incoming data stream 
			\State Train the CZSL model on $\mathcal{M}$ for multiple epochs on the available data in replay memory $\mathcal{M}$.
			\State $optimization\_done$ $\gets$ True
			\Else
			\If{$optimization\_done$ is True} 
			\State Store $i^{th}$ sample in a short-term batch memory $\mathcal{M}_b$ 
			\If{short-term batch memory $\mathcal{M}_b$ is full}
			\State Train the CZSL model continuously on the incoming batch of samples available in $\mathcal{M}_b$ and samples taken from replay memory $\mathcal{M}$ without running any epochs.
			\State Clear the short-term batch memory $\mathcal{M}_b$
			\Else
			\State Keep model in sleep for very small duration of time 
			\EndIf
			\EndIf
			\EndIf
			\EndFor
		\end{algorithmic}
		\label{alg:ALG1}
	\end{algorithm}
	
	\begin{algorithm}[t]
		\caption{{\bf Task-free Learning Strategy-2}}
		\textbf{Input:} \text{Data stream $\mathcal{D}_{tr}$, Short-term memory $\mathcal{M}_{st}$}\\
		\textbf{Output:} \text{Trained model}
		\begin{algorithmic}[1]
			\For{$i^{th}$ sample in $\mathcal{D}_{tr}$}
			\State Store the incoming $i^{th}$ sample in replay memory $\mathcal{M}$ using Reservoir sampling strategey
			\State Store the incoming $i^{th}$ sample in the short-term memory $\mathcal{M}_{st}$
			\If{$M_{st}$ is full}
			\State Train the CZSL model on the samples taken from replay memory $\mathcal{M}$ and short-term memory $\mathcal{M}_{st}$ for multiple  epochs to optimize the parameters
			\State Clear the short-term memory $\mathcal{M}_{st}$
			\Else
			\State Keep model in sleep until $\mathcal{M}_{st}$ is not full 
			\EndIf
			\EndFor
		\end{algorithmic}
		\label{alg:ALG2}
	\end{algorithm}
	
	\subsection{Knowledge Distillation (KD) Using Dark Knowledge for CZSL}
	In addition to ER, Tf-GCZSL also performs KD with dark knowledge for mitigating catastrophic forgetting of the model. For this purpose, in addition to storing the training sample in $\mathcal{M}$,  class attribute information and latent space information (i.e., estimated $\mu_{Vf}$, $\Sigma_{Vf}$ $\mu_{Af}$, and $\Sigma_{Af}$ by the encoder) corresponding to the training sample are also stored. These latent space information is dark knowledge, which is used to perform knowledge distillation using dark knowledge ($\mathcal{L}^{dark}_{KD}$) as:
	\begin{equation}
	\mathcal{L}^{dark}_{KD} = \|\mu_{Af} - \mu_{Af_\mathcal{M}}\|_1 + \|\Sigma_{Vf} - \Sigma_{Vf_\mathcal{M}}\|_1,
	\end{equation}     
	where $\mu_{Af_\mathcal{M}}$ and $\Sigma_{Vf_\mathcal{M}}$ are retrieved from the stored latent information for the corresponding sample in $\mathcal{M}$. These values were estimated by an encoder at any point of time in the past on the learning trajectory of the Tf-GCZSL. One should note that the approach does not store/use any previously trained network as a teacher for performing knowledge distillation. Instead the knowledge required to perform distillation is stored into the $\mathcal{M}$ with sample information.  
	
	\subsection{Overall training procedure of Tf-GCZSL:} 
	
	Overall, Tf-GCZSL minimizes the following loss during training:
	\begin{equation}\label{tfgzsl_loss}
	\mathcal{L}_{G} = \mathcal{L}_{Re} + \beta\mathcal{L}_{KL} + \gamma \mathcal{L}_{CA} + \delta \mathcal{L}_{DA} + \alpha \mathcal{L}^{dark}_{KD},
	\end{equation}
	where $\beta$, $\gamma$, $\delta$, and $\alpha$ are the weighting factors. For the task-free CZSL, first minimize the loss and follow one of the two above-discussed task-free training strategies, i.e., either Tf-GCZSL$_{\mathcal{M}_{b}}$ or Tf-GCZSL$_{\mathcal{M}_{st}}$. 
	
	After completion of training, latent features are generated by sampling based on the estimated mean and variance of visual/attribute encoder. Visual encoder is used to generate latent features for seen classes and attribute encoder is used for unseen class. Since, these latent features are very discriminative, a simple linear classifier using Softmax is trained on these latent features. The proposed Tf-GCZSL method can also used for the task-agnostic prediction where task boundary known at training time but not at testing time. In this case, Tf-GCZSL minimize the same loss function with-out the task-free learning strategy.

	%
	%	\subsection{Class Incremental Learning}\label{subsec:class_inc}
	%We are solving a CZSL problem for single-head setting as it provides a task-agnostic prediction, as well as task-free learning. If all data are available for training, then any classifier can learn all of these classes jointly. However, for the CZSL problem, data comes sequentially in the form of a task. We generate latent features using $t^{th}$ encoder of Tf-GCZSL for $t^{th}$ task and use these latent features for classification. These latent features are beneficial in two ways: (a) we can generate as many samples we want for underrepresented classes; (b) these latent features are quite discriminative and can be easily classified. Since latent features are discriminative, a simple linear classifier with softmax is used for classification. For a new class at $t^{th}$ task, a trained linear classifier of $(t-1)^{th}$ task can be easily and efficiently extended by extending softmax as mentioned in \cite{liu2020generative}. 

	\section{Performance Evaluation}
	
	CZSL methods have been evaluated over five benchmark ZSL datasets, namely Caltech-UCSD-Birds 200-2011 (CUB) \cite{CUB}, Attribute Pascal and Yahoo (aPY) \cite{aPY}, Animals with Attributes (AWA1 and AWA2) \cite{aPY}, and SUN \cite{SUN}. These datasets split and prepared for two kinds of CZSL settings, which is discussed in the next subsection.
	
	\subsection{Settings and Evaluation Metrics}
	In this section, we discuss two different settings  for evaluating our proposed model in task-agnostic prediction and task-free learning for continual zero-shot learning (CZSL) .  
	
	%	\myparagraph{CZSL setting for task-agnostic prediction:}
	%	For task-agnostic prediction, we have used the setting mentioned in \cite{skorokhodov2020normalization} for CZSL. It is briefly provided in supplementary material. 
	
	{
		\renewcommand{\arraystretch}{1.4} 
		\begin{table*}[t]
			\centering
			\resizebox{\textwidth}{!}{%
				\begin{tabular}{|l|ccc|ccc|ccc|ccc|ccc|}
					\hline
					& \multicolumn{3}{c|}{CUB} & \multicolumn{3}{c|}{aPY} & \multicolumn{3}{c|}{AWA1} & \multicolumn{3}{c|}{AWA2} & \multicolumn{3}{c|}{SUN} \bigstrut[b]\\
					\cline{2-16}          & \multicolumn{1}{p{4.215em}}{mSA} & \multicolumn{1}{p{2.215em}}{mUA} & \multicolumn{1}{p{2.215em}|}{mH} & \multicolumn{1}{p{2.215em}}{mSA} & \multicolumn{1}{p{2.215em}}{mUA} & \multicolumn{1}{p{2.215em}|}{mH} & \multicolumn{1}{p{2.215em}}{mSA} & \multicolumn{1}{p{2.215em}}{mUA} & \multicolumn{1}{p{2.215em}|}{mH} & \multicolumn{1}{p{2.215em}}{mSA} & \multicolumn{1}{p{2.215em}}{mUA} & \multicolumn{1}{p{2.215em}|}{mH} & \multicolumn{1}{p{2.215em}}{mSA} & \multicolumn{1}{p{2.215em}}{mUA} & \multicolumn{1}{p{2.215em}|}{mH} \bigstrut\\
					\hline
					Seq-Tf-GCZSL & 40.82 & 14.37 & 21.14 & 47.00 & 7.83  & 13.13 & 50.81 & 16.68 & 25.45 & 52.24 & 13.98 & 22.33 & 25.94 & 16.22 & 20.10 \bigstrut[t]\\
					AGEM+CZSL \cite{chaudhry2018efficient} & --  & --  & 17.30 & --  & --  & --  & --  & --  & --  & --  & --  & --  & --  & --  & 9.60 \\
					AGEM+CZSL+CN \cite{skorokhodov2020normalization} & --  & --  & 23.80 & --  & --  & --  & --  & --  & --  & --  & --  & --  & --  & --  & 14.20 \\
					EWC+CZSL \cite{schwarz2018progress} & --  & --  & 18.00 & --  & --  & --  & --  & --  & --  & --  & --  & --  & --  & --  & 9.60 \\
					EWC+CZSL+CN \cite{skorokhodov2020normalization} & --  & --  & 23.30 & --  & --  & --  & --  & --  & --  & --  & --  & --  & --  & --  & 14.30 \\	
					MAS+CZSL \cite{aljundi2018memory} & --  & --  & 17.70 & --  & --  & --  & --  & --  & --  & --  & --  & --  & --  & --  & 9.40 \\														
					MAS+CZSL+CN \cite{skorokhodov2020normalization} & --  & --  & 23.80 & --  & --  & --  & --  & --  & --  & --  & --  & --  & --  & --  & 14.20 \\ \hline
					Tf-GCZSL$_{NK}$ & 45.00 & 30.50 & 34.57 & 58.41 & 18.74 & 26.85 & 61.67 & 37.38 & 44.90 & 65.46 & 36.40 & 45.75 & 27.07 & 23.35 & 23.84 \\
					Tf-GCZSL & 46.63 & 32.42 & \textbf{36.31} & 57.92 & 21.22 & \textbf{29.55} & 64.00 & 38.34 & \textbf{46.14} & 64.89 & 40.23 & \textbf{48.33} & 28.09 & 24.70 & \textbf{24.79} \bigstrut[b]\\
					\hline
				\end{tabular}%
			}
			\vspace{1mm}
			\caption{CZSL results for task-agnostic prediction in terms of mean seen accuracy (mSA) for seen, mean unseen accuracy (mUA) for unseen classes, and their mean of harmonic mean (mH). The best results in the table are presented in bold face.}
			\label{tab:gen_res_S1}%
		\end{table*}%
	}

	{
		\renewcommand{\arraystretch}{1.4} 
		\begin{table*}[t]
			\centering
			\resizebox{\textwidth}{!}{%
				\begin{tabular}{|l|ccc|ccc|ccc|ccc|ccc|}
					\hline
					& \multicolumn{3}{c|}{CUB} & \multicolumn{3}{c|}{aPY} & \multicolumn{3}{c|}{AWA1} & \multicolumn{3}{c|}{AWA2} & \multicolumn{3}{c|}{SUN} \bigstrut[b]\\
					\cline{2-16}          & \multicolumn{1}{p{4.215em}}{SA} & \multicolumn{1}{p{2.215em}}{UA} & \multicolumn{1}{p{2.215em}|}{H} & \multicolumn{1}{p{2.215em}}{SA} & \multicolumn{1}{p{2.215em}}{UA} & \multicolumn{1}{p{2.215em}|}{H} & \multicolumn{1}{p{2.215em}}{SA} & \multicolumn{1}{p{2.215em}}{UA} & \multicolumn{1}{p{2.215em}|}{H} & \multicolumn{1}{p{2.215em}}{SA} & \multicolumn{1}{p{2.215em}}{UA} & \multicolumn{1}{p{2.215em}|}{H} & \multicolumn{1}{p{2.215em}}{SA} & \multicolumn{1}{p{2.215em}}{UA} & \multicolumn{1}{p{2.215em}|}{H} \bigstrut\\
					
					\hline
					\multicolumn{1}{|c|}{Offline (Upper Bound)} & 53.5  & 51.6  & 52.4  & 59.36 & 30.36 & 40.18 & 72.8  & 57.3  & 64.1  & 75    & 55.8  & 63.9  & 35.7  & 47.2  & 40.6 \bigstrut\\  \hline
					Seq-Tf-GCZSL$_{\mathcal{M}_{b}}$ & 38.73 & 21.42 & 27.58 & 23.51 & 1.68  & 3.14  & 20.37 & 2.12  & 3.85  & 14.64 & 9.43  & 11.47 & 27.89 & 18.88 & 22.52 \bigstrut[t]\\
					Seq-Tf-GCZSL$_{\mathcal{M}_{st}}$ & 42.48 & 20.46 & 27.61 & 57.23 & 3.90  & 7.30  & 55.00 & 13.99 & 22.31 & 59.67 & 18.37 & 28.09 & 26.42 & 17.84 & 21.30 \bigstrut[t]\\ \hline
					Tf-GCZSL$_{\mathcal{M}_{b}-NK}$ & 45.69 & 31.90 & 37.57 & 73.13 & 15.51 & 25.60 & 67.65 & 44.01 & 53.33 & 70.08 & 46.59 & 55.97 & 31.04 & 29.30 & 30.15 \bigstrut[t]\\
					Tf-GCZSL$_{\mathcal{M}_{st}-NK}$ & 46.70 & 43.09 & 44.82 & 77.68 & 16.67 & 27.46 & 66.24 & 53.28 & 59.06 & 69.38 & 55.07 & 61.40 & 26.86 & 37.56 & 31.32 \bigstrut[t]\\ \hline					
					Tf-GCZSL$_{\mathcal{M}_{b}}$ & 45.08 & 34.02 & 38.78 & 72.55 & 14.33 & 23.94 & 65.64 & 51.46 & 57.69 & 68.42 & 42.74 & 52.62 & 31.00 & 29.37 & 30.16 \bigstrut[t]\\
					Tf-GCZSL$_{\mathcal{M}_{st}}$ & 44.52 & 43.21 & 43.85 & 72.12 & 19.66 & 30.90 & 61.79 & 57.77 & 59.72 & 67.42 & 58.08 & 62.41 & 27.76 & 39.09 & 32.46 \bigstrut[b]\\
					\hline
				\end{tabular}%
			}
			\vspace{1mm}
			\caption{Results for task-free CZSL in terms of seen accuracy (mSA) for seen, unseen accuracy (UA) for unseen classes, and their harmonic mean (H) .}
			\label{tab:gen_res_S2}%
		\end{table*}%
	}	
	
	\myparagraph{CZSL setting for task-agnostic prediction:}
	For task-agnostic prediction, we have used the setting mentioned in \cite{skorokhodov2020normalization} for CZSL. In this setting, first, data is divided among $T$ tasks. Next, if the model is training on $t^{th}$ task then all classes till $t^{th}$ task are treated as seen classes and all classes from $(t+1)^{th}$ task to $T$ tasks are treated as unseen classes. Following evaluation metrics are used to evaluate the model in case of task-agnostic prediction for a $t^{th}$ task \cite{skorokhodov2020normalization}:
	\begin{itemize}%\itemsep-0.5em
		\item Mean Seen-class Accuracy (mSA)
		\begin{equation}
		mSA = \frac{1}{T}\sum_{t=1}^T CAcc(\mathcal{D}_{ts}^{\leq t}, A^{\leq t}),
		\end{equation}
		where $CAcc$ stands for per class accuracy. 
		\item Mean Unseen-class Accuracy(mUA)
		\begin{equation}
		mUA = \frac{1}{T-1}\sum_{t=1}^{T-1} CAcc(\mathcal{D}_{ts}^{> t}, A^{> t})
		\end{equation}
		\item Mean Harmonic Accuracy (mH)
		\begin{equation}
		mH = \frac{1}{T-1}\sum_{t=1}^{T-1} H(\mathcal{D}_{ts}^{\leq t}, \mathcal{D}_{ts}^{> t}, A),
		\end{equation}
		where $H$ stands for harmonic mean.
		%		\item Mean Joint Accuracy(mJA)
		%		\begin{equation}
		%		mJA = \frac{1}{T}\sum_{t=1}^{T} Acc(D_{ts}, A)
		%		\end{equation}
		
	\end{itemize}
	
	Here, $\mathcal{D}^{\leq t}$ denotes all train/test samples from $1^{st}$ to $t^{th}$ task, and
	$D^{> t}$ denotes all train/test samples from $(t+1)^{th}$ to last task.
	
	\myparagraph{CZSL setting for task-free learning:} The setting  mentioned above in \cite{skorokhodov2020normalization} is not suitable for task-free GCZSL as seen and unseen classes are decided based on the task boundary. However, in task-free learning, task boundary information is not available during training and testing of the model. Therefore, we propose a more challenging and different CZSL setting for task-free learning. Here data split into multiple blocks based on the standard split of ZSL benchmark datasets. Each block contains samples from distinct classes. First, we train the model by streaming sample from these blocks one by one, then, test the model on the standard testing data available in the split of ZSL benchmark datasets. The performance is evaluated using the harmonic mean ($H$), and top-1 accuracy of seen-class accuracy (SA) and unseen classes accuracy (UA).    
	
	\subsection{Baseline Methods}
	There are only handful of work available for CZSL. Recently, it is developed for multi-head setting \cite{wei2020lifelong} and task-agnostic prediction \cite{skorokhodov2020normalization}; however, there is no work available for task-free learning. For task-agnostic prediction, the results are compared with the following methods: 
	\begin{itemize}
		\item The sequential training of the proposed method without considering any continual learning setting: Seq-Tf-GCZSL. 
		\item Skorokhodov et al. developed various methods for CZSL with and without class normalization \cite{skorokhodov2020normalization}: 
		\begin{enumerate}[(i)]
			\item With class normalization: AGEM+CZSL \cite{chaudhry2018efficient}, EWC+CZSL \cite{schwarz2018progress}, MAS+CZSL \cite{aljundi2018memory}.
			\item Without class normalization: AGEM+CZSL+CL, EWC+CZSL+CL, MAS+CZSL+CL
		\end{enumerate}
	\end{itemize}
	For task-free learning, the results are compared with the offline training of the proposed method where one can assume all data are available at once, which is basically an upper bound for the proposed method. We also perform sequential training of the proposed methods: Seq-Tf-GCZSL$_{\mathcal{M}_{b}}$ and Seq-Tf-GCZSL$_{\mathcal{M}_{st}}$.

	\begin{figure*}[!h]
		\centering
		\begin{subfigure}{.43\textwidth}
			\centering
			\includegraphics[width=1.1\linewidth,height=4.0cm]{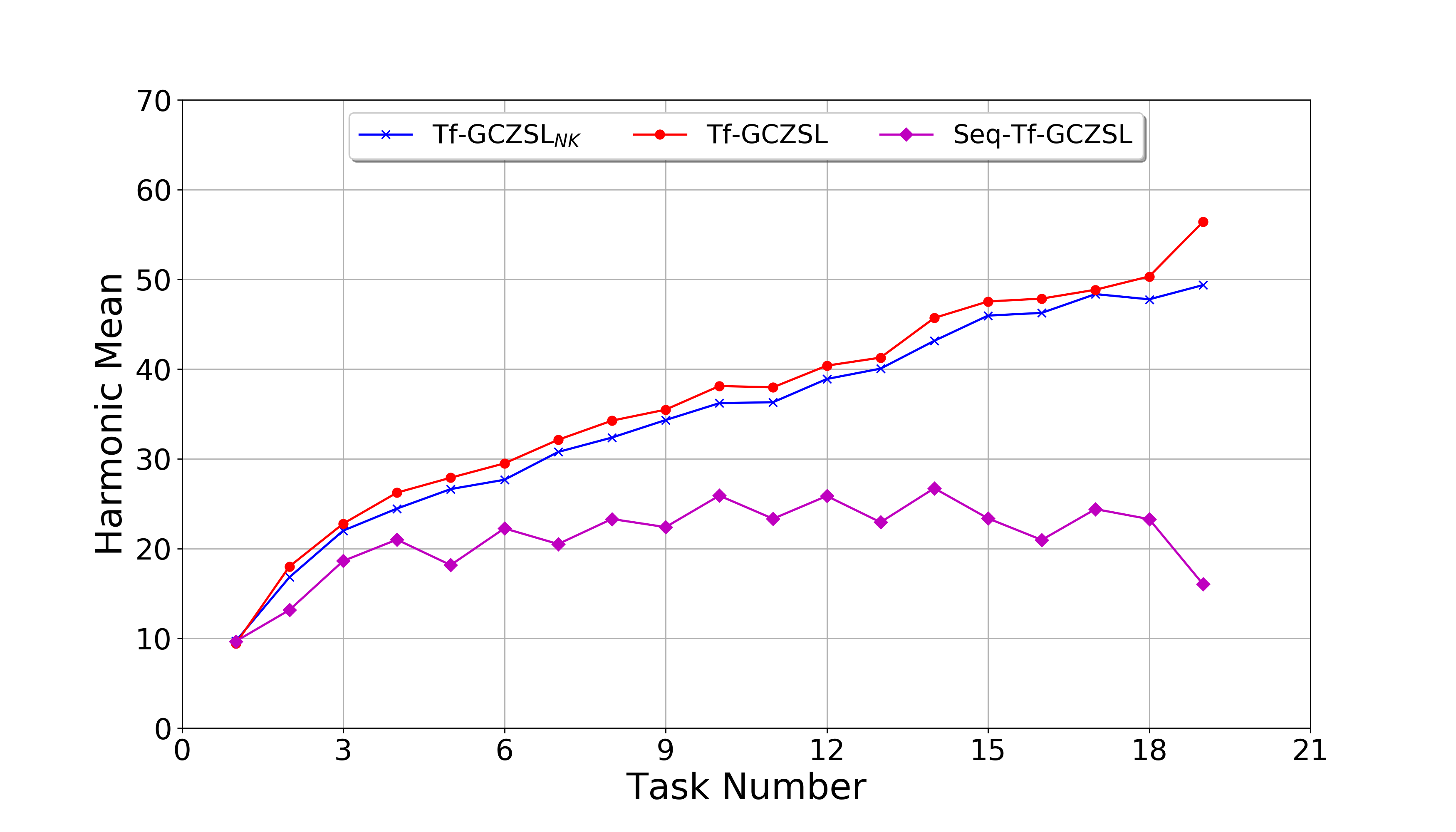}
			\caption{Task-wise analysis}
			\label{fig:cub-mH_s1}
		\end{subfigure}
		\begin{subfigure}{.28\textwidth}
			\centering
			\includegraphics[width=1.1\linewidth,height=4.0cm]{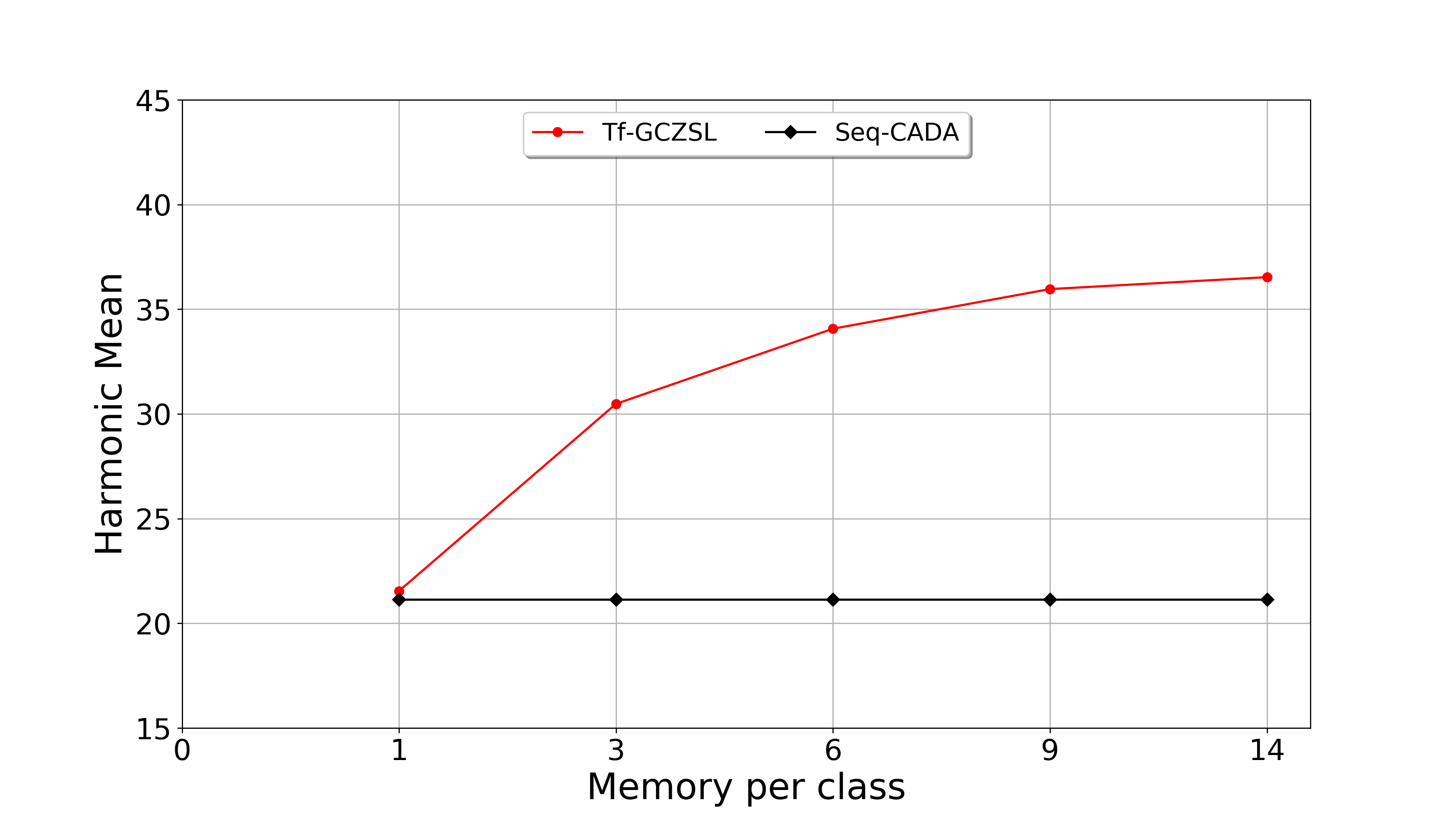}
			\caption{Impact of the size of the memory $\mathcal{M}$, which is analyzed in terms of number of samples per class.}
			\label{fig:cub-mem_s1}
		\end{subfigure}
		\begin{subfigure}{.28\textwidth}
			\centering
			\includegraphics[width=1.1\linewidth,height=4.0cm]{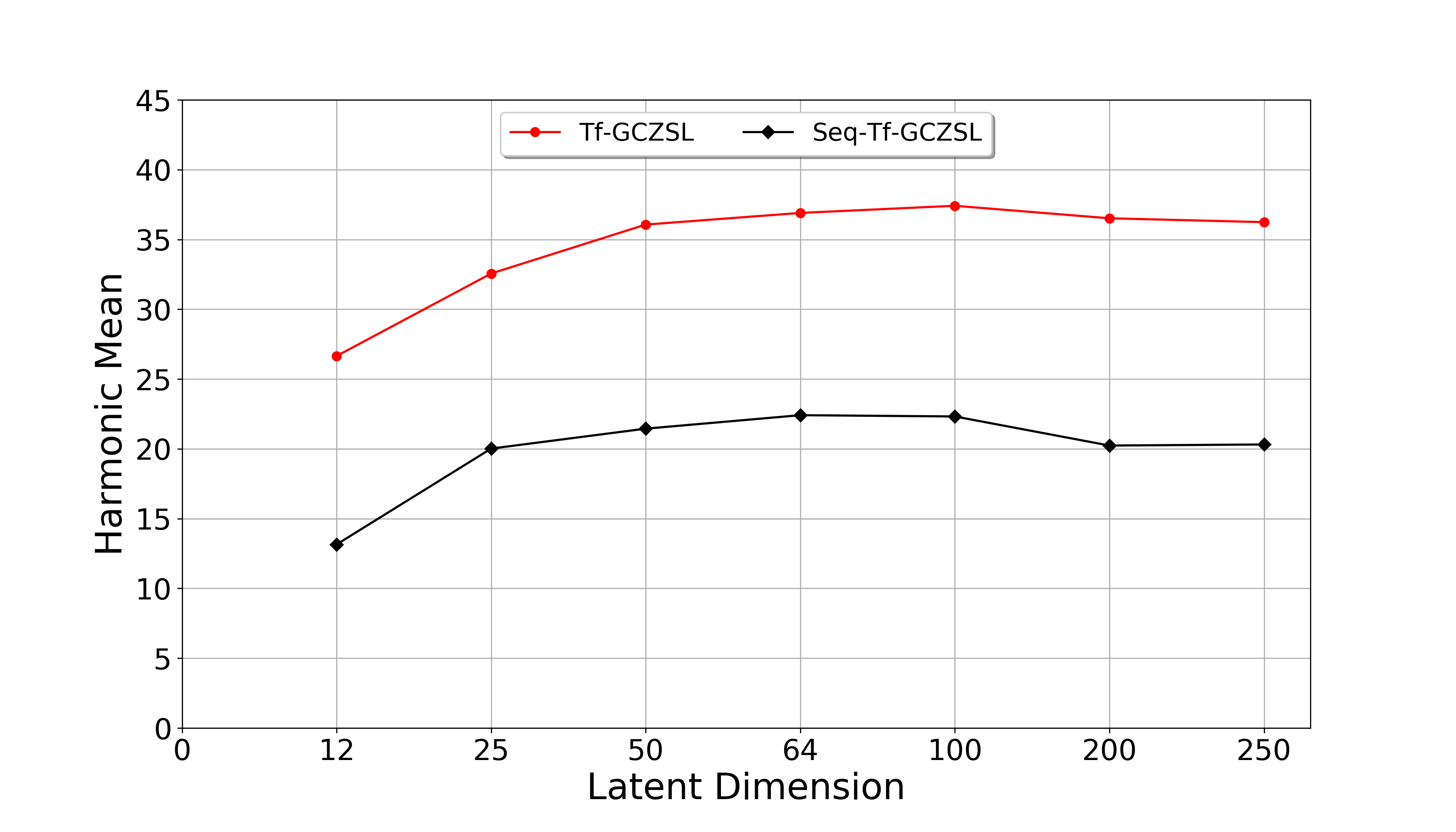}
			\caption{Impact of the latent dimensions}
			\label{fig:cub-lat_s1}
		\end{subfigure}	
		\caption{Ablation study for task-agnostic prediction}
		\label{fig:taskwise-mH}
	\end{figure*}
	
	\begin{figure*}[!h]
		\centering
		\begin{subfigure}{.33\textwidth}
			\centering
			\includegraphics[width=1.1\linewidth,height=4.0cm]{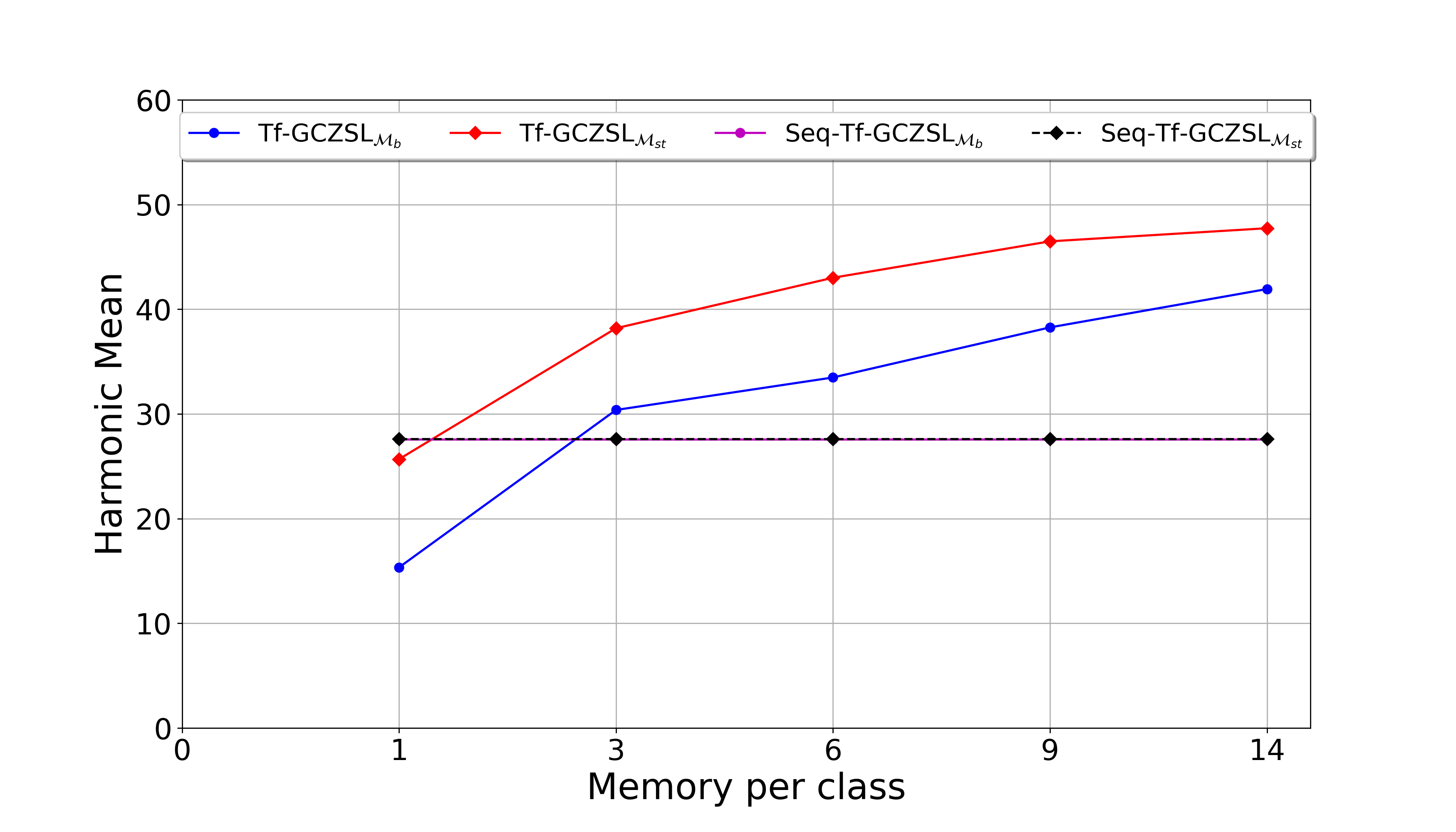}
			\caption{Impact of the size of the memory $\mathcal{M}$, which is analyzed in terms of number of samples per class.}
			\label{fig:cub-mem_s2}
		\end{subfigure}
		\begin{subfigure}{.33\textwidth}
			\centering
			\includegraphics[width=1.1\linewidth,height=4.0cm]{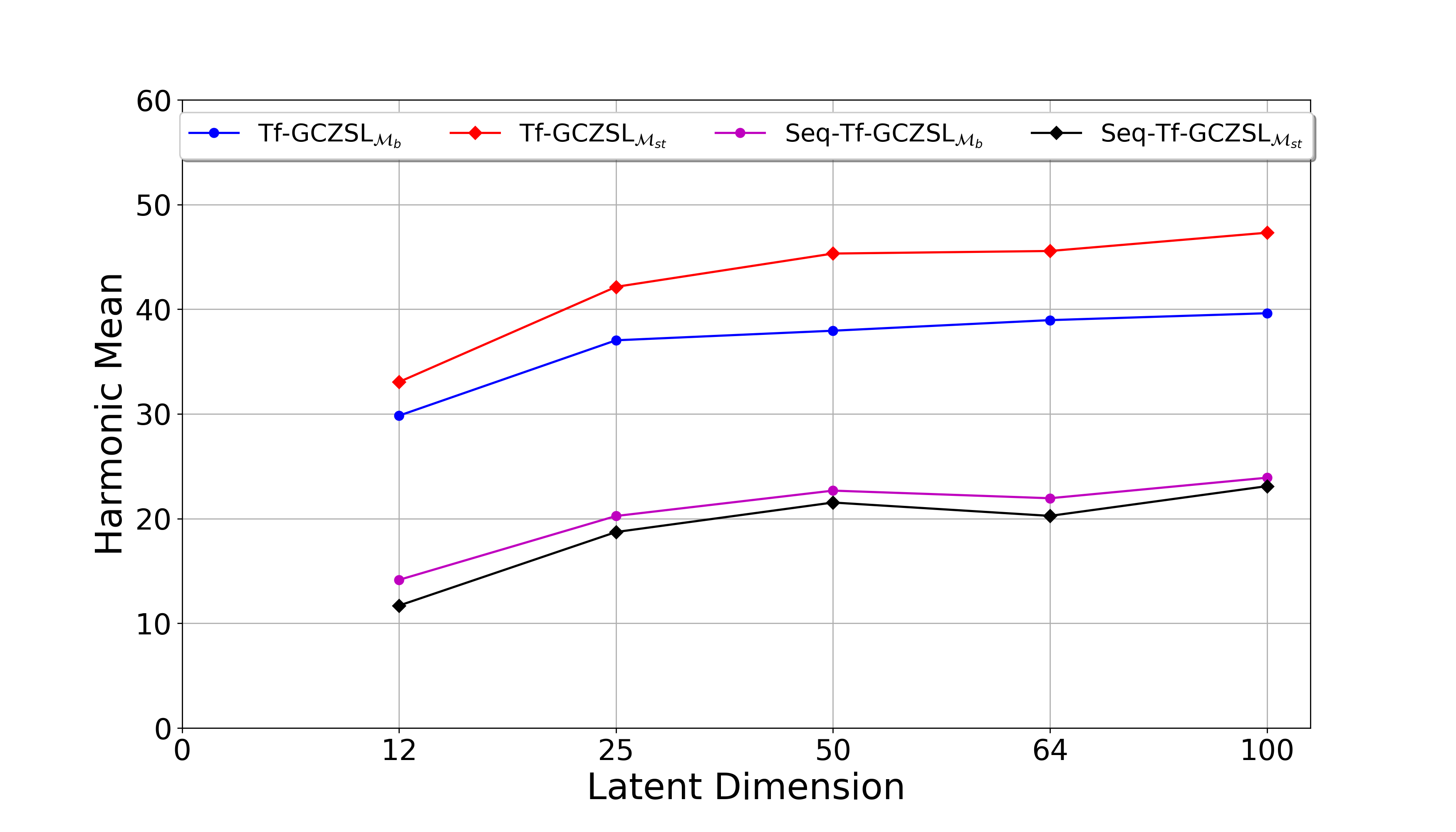}
			\caption{Impact of the latent dimensions}
			\label{fig:cub-lat_s2}
		\end{subfigure}
		\begin{subfigure}{.33\textwidth}
			\centering
			\includegraphics[width=1.1\linewidth,height=4.0cm]{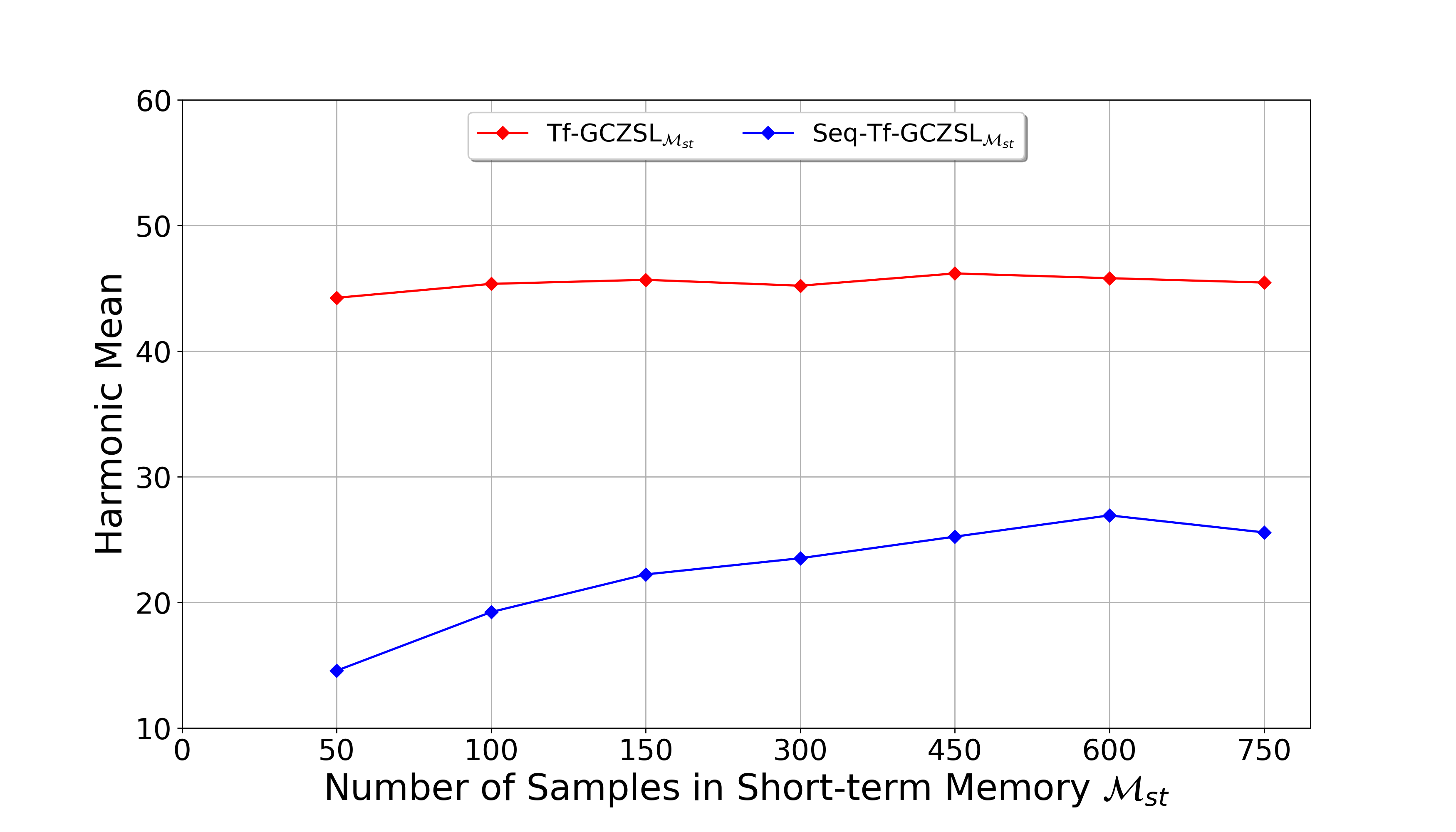}
			\caption{Impact of the size of $_{\mathcal{M}_{st}}$}
			\label{fig:cub-stm_s2}
		\end{subfigure}	
		\caption{Ablation study for task-free prediction}
		\label{fig:taskfree-mH}
	\end{figure*}

	\subsection{Results}
	
	In this section, results are presented for the both cases of single head setting, i.e. task-agnostic prediction and task free learning.
	
	\myparagraph{Task-agnostic Prediction:}
	Results for this setting are presented in Table \ref{tab:gen_res_S1} for 5 CZSL datasets. The performance results of Tf-GCZSL without KD using dark knowledge are given in the table, i.e., Tf-GCZSL$_{NK}$. It can be observed from this table, Tf-GCZSL outperforms the all existing CZSL methods presented in \cite{skorokhodov2020normalization} from at least $12\%$ and $10\%$ margin for CUB and SUN datasets in terms of $mH$, respectively. It has also significantly outperform the seq-Tf-GCZSL, which is obvious. Moreover, when we compare the results of Tf-GCZSL and Tf-GCZSL$_{NK}$, it has been observed that KD using dark knowledge with experience replay improves the performance and helps in alleviating the catastrophic forgetting further.    
	
	%	Further, we also provide the results of the proposed methods without dark knowledge-based KD: Tf-GCZSL$_{\mathcal{M}_{b}-NK}$ and Tf-GCZSL$_{\mathcal{M}_{st}}$      
	
	\myparagraph{Task Free Learning:}  
	Results for this setting are presented in Table \ref{tab:gen_res_S2} for all 5 CZSL datasets. We also provide the results of the proposed methods in the sequential setting (i.e., Seq-Tf-GCZSL$_{\mathcal{M}_{b}}$ and Seq-Tf-GCZSL$_{\mathcal{M}_{st}}$), and without KD using dark knowledge (i.e., Seq-Tf-GCZSL$_{\mathcal{M}_{b}-NK}$ and Seq-Tf-GCZSL$_{\mathcal{M}_{st}-NK}$).	It can be observed from this table, proposed methods outperforms sequential methods for all datasets. Further, dark knowledge improves the performance for most of the case if we use $\mathcal{M}_{st}$. Overall, in the second strategy-based task-free learning (Tf-GCZSL$_{\mathcal{M}_{st}}$) outperforms significantly over the first strategy (i.e., Tf-GCZSL$_{\mathcal{M}_{b}}$) by more than $5\%$, $6\%$, $2\%$, $9\%$, and $2\%$ for CUB, aPY, AWA1, AWA2, and SUN datasets, respectively in terms of $H$. Moreover, When we compare from the upper bound, Tf-GCZSL$_{\mathcal{M}_{st}}$ yields good performance as it lacks by only $8.55\%$, $9.28\%$, $4.38\%$, $1.49\%$, and $8.14\%$ for CUB, aPY, AWA1, AWA2, and SUN datasets, respectively in terms of $H$. This lack of performance is due to catastrophic forgetting. 
	
	Tf-GCZSL gains this performance for both settings due to the joint training of the current samples with the samples from the replay memory $\mathcal{M}$. Although, replay memory does the repetitive training of the already trained samples, but it does not lead to overfitting or any adverse effect. In contrast, this joint training regularize the model for better generalization.   
	
	\subsection{Ablation Study on CUB dataset}
	
	In this section, an ablation study is presented for the CUB dataset.
	
	\myparagraph{For task-agnostic prediction}: The ablation study is presented in terms of the following three factors: 
	\begin{itemize}
		\item Task-wise analysis: Task-wise analysis is depicted in Figure \ref{fig:cub-mH_s1}. It can be observed from this figure, performance of Tf-GCZSL improves as number of tasks increases because as number of task increases then task-relatedness between seen and unseen class samples are increasing. Task-relatedness increases because number of seen samples  also increases and it enriches the knowledge of the model. Although Tf-GCZSL improves the performance when task increase, however, seq-Tf-GCZSL performance has been decreased as it doesn't use any continual learning strategy.  
		
		\item Analysis on replay memory: The performance of Tf-GCZSL is very sensitive to the size of the memory. Size is kept in term of number of samples per class. If there are $j$ number of classes and $k$ number of samples per class then memory size is $j*k$. It can be observed from Figure \ref{fig:cub-mem_s1}, the performance of TF-GCZSL improves as memory size increases as memory can keep more number of samples from the past experience. 
		
		\item Analysis on latent dimensions: It is another important factor for CZSL. Performance of TF-GCZSL is depicted in \ref{fig:cub-lat_s1} on different latent dimensions. This figures suggests that size of latent dimensions should not be very small or very large. If it is very small then it is unable to create the more discriminative feature and if it is very large then degree of freedom increases which will not provide the compact features.              
	\end{itemize}
	
	\myparagraph{For task-free learning:} Similarly for task-agnostic prediction,  the ablation study is conducted on memory size and latent dimensions. Since, Tf-GCZSL uses two kinds of memories: replay memory $\mathcal{M}$ and short-term memory $\mathcal{M}_{st}$. Analysis-based on both memories is presented in Figures \ref{fig:cub-mem_s2} and \ref{fig:cub-stm_s2} for $\mathcal{M}$ and $\mathcal{M}_{st}$, respectively. In case of $\mathcal{M}$, performance increases as memory size increases due to the same reason as discussed above. In case of $\mathcal{M}_{st}$, size is not impacting much on the performance as these samples jointly trains with the larger memory $\mathcal{M}$ in the task-free setting, therefore, performance is very similar for all cases.  In the ablation of latent dimension in Figure \ref{fig:cub-lat_s2}, we again observe the similar plot as \ref{fig:cub-lat_s1}. Moreover, for all three plots in Figure \ref{fig:taskfree-mH}, we also plot sequential results for better understanding and results are obvious that proposed methods outperforms sequential methods for all cases.

	\section{Conclusion}
	This is the first work that tackles the continual Zero-shot learning for the task-free set-up to the best of our knowledge. This paper has proposed general task-free continual zero-shot learning strategies using experience replay, knowledge distillation with dark knowledge, and short-term memory. The performance is evaluated on five benchmark data, and the results indicate that the Tf-GCSZL achieve closer to the upper bound with minimal catastrophic forgetting. The framework is generic; therefore, one can use other ZSL approaches to develop it for task-free CZSL.

	{\small
		\bibliographystyle{ieee_fullname}
		\bibliography{CZSL}
	}
	
\end{document}